\documentclass[11pt,en,bibtex,bibend=bibtex]{elegantpaper}

\usepackage{fancyhdr}
\usepackage{graphicx}
\usepackage{float}
\usepackage{amsfonts} % 用于 \mathbb{I}

\newcommand{\logoheight}{0.8cm}   % Logo 高度
\newcommand{\logofile}{logo_text.pdf}  % 请确保文件夹下有此文件
\fancypagestyle{plain}{
    \fancyhf{} % 清空默认
    \fancyhead[L]{\includegraphics[height=\logoheight]{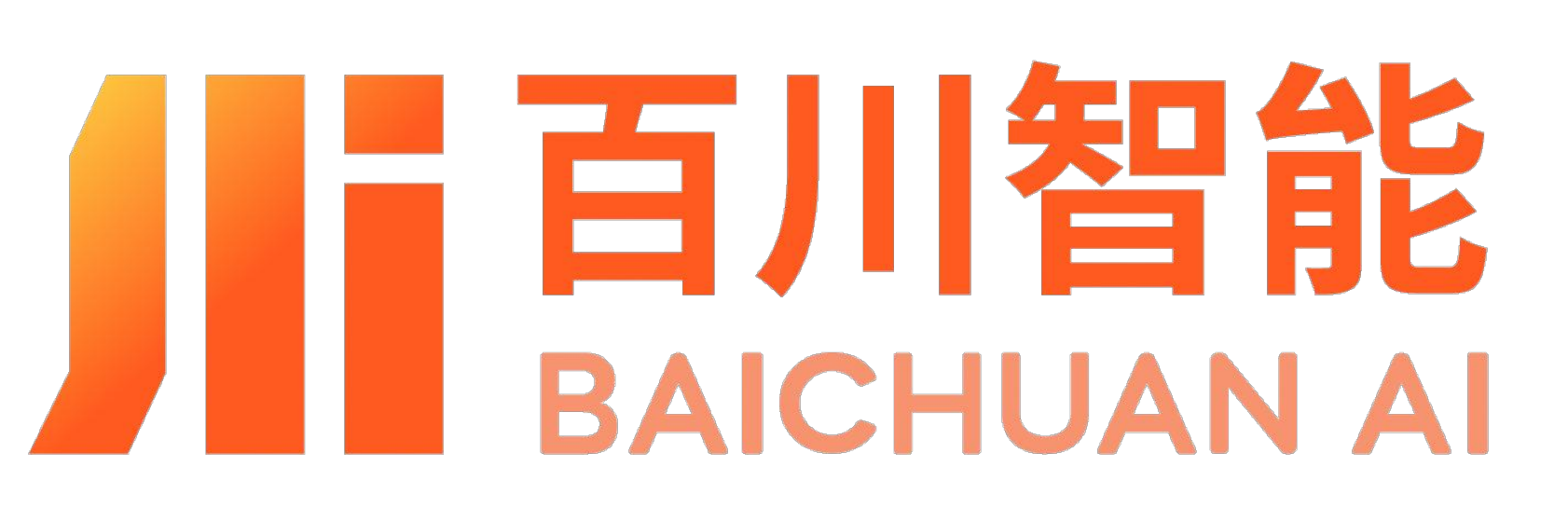}}
    \fancyhead[R]{\footnotesize\itshape Baichuan-M4}
    \fancyfoot[C]{\thepage}
}

\fancypagestyle{fancy}{
    \fancyhf{} % 清空默认
    \fancyhead[L]{}
    \fancyhead[R]{\footnotesize\itshape Baichuan-M4 Model Card}
    \fancyfoot[C]{\thepage}
}

\title{Baichuan-M4: A Clinical-Grade Medical Agent System for Continuous Care}

\author{%
\textbf{Baichuan AI and THUBPM Group, Tsinghua University}\\[0.7em]
\normalsize\normalfont
See the \hyperref[sec:contribution]{Contribution} section for a full author list.%
}

\usepackage{array}

\usepackage{tabularx}
\newcolumntype{C}{>{\centering\arraybackslash}X}

\AtBeginBibliography{\sloppy\emergencystretch=1em\raggedright}

\setlist{topsep=4pt, partopsep=0pt, itemsep=2pt, parsep=0pt}
\begin{document}

% Slightly relax line breaking to avoid minor overfull boxes
\setlength{\emergencystretch}{1em}

\maketitle

\begin{abstract}
\noindent
Baichuan-M4 is Baichuan Intelligence's clinical-grade medical large language model, designed for physician-supervised \emph{continuous care} rather than single-turn medical question answering. It is built as a coordinated medical agent system around three pillars: \textbf{Baichuan-Harness}, a unified runtime that keeps reinforcement-learning training and real-world deployment consistent while enforcing action constraints, tool use, long-term patient memory, and multi-agent coordination; a \textbf{core reasoning model} trained with a continuous-care reinforcement-learning framework that integrates span-level reward modeling (SPAR++), reasoning-path compression, curriculum learning, and stabilized policy optimization; and a \textbf{clinical tool layer} for patient-memory management, authoritative evidence-based retrieval, and multimodal medical perception across documents, X-rays, and dermatology. On a cross-dimensional medical evaluation suite, Baichuan-M4 attains leading results in static medical knowledge and safety, dynamic OSCE-style consultation, long-context clinical memory, evidence-based retrieval, medical document OCR, and multimodal image understanding, while lowering the hallucination rate to 3.3\%.
\end{abstract}


% [中文] 一、模型总览
\section{Model Overview}

% [中文] Baichuan-M4 是百川智能新一代临床级医疗大模型，面向连续照护场景打造。相比已经具备多轮医学问诊能力、聚焦于模型自身医学对话与专业可用性的 Baichuan-M3，Baichuan-M4 进一步面向真实医疗场景中长期、多阶段、跨模态的诊疗需求设计，系统性引入长期患者记忆、工具调用、循证检索与 Agent 编排机制，构建了以 Harness、Model、Tool 为核心的医疗 Agent 系统。
\textbf{Baichuan-M4} is Baichuan Intelligence's next-generation clinical-grade medical LLM~\cite{bcm1, baichuan-m2}, developed for physician-supervised \textbf{continuous care scenarios}. Compared with Baichuan-M3~\cite{m3team2026baichuanm3modelingclinicalinquiry}, which already supported multi-turn medical consultation and focused on medically reliable dialogue capabilities, Baichuan-M4 is further designed for the long-term, multi-stage, and cross-modal diagnostic and treatment needs found in controlled real-world healthcare settings. It systematically incorporates long-term patient memory, tool use, evidence-based retrieval, and agent orchestration, building a medical agent system centered on Harness, Model, and Tool.

% [中文] Baichuan-M4 的目标不只是一个医学知识问答模型。它旨在作为一个临床级医疗 Agent 系统，能够在受控环境中完成患者问诊、随访、持续照护、循证检索、医学影像理解、长期患者记忆管理与多 Agent 协同。
Baichuan-M4 is designed as a \textbf{clinical-grade medical agent system}, supporting patient consultation, follow-up, continuous care, evidence-based retrieval, medical image understanding, long-term patient memory, and multi-agent coordination in controlled environments.

\paragraph{Definition of clinical-grade.}
In this white paper, ``clinical-grade'' refers to readiness for physician-supervised use in controlled deployment settings, under predefined requirements for evaluation, safety, evidence traceability, and clinical workflow constraints. It does not imply autonomous diagnosis, prescription, or regulatory approval as a medical device.

% [中文] 1.1 核心能力升级
\subsection{Core Capability Upgrades}

% [中文] 下表总结了从 Baichuan-M3 到 Baichuan-M4 的核心能力升级。
Table~\ref{tab:capability-upgrades} summarizes the core capability upgrades from Baichuan-M3 to Baichuan-M4.

% [中文] 表1：从 Baichuan-M3 到 Baichuan-M4 的核心能力升级。
% [中文] 表头：维度 | Baichuan-M3 | Baichuan-M4
% [中文] 问诊能力：具备多轮医学问诊能力，聚焦单次就诊场景中的专业医学对话 → 面向长期、多阶段连续照护的动态问诊与随访
% [中文] 循证能力：具备医学知识推理与经验性判断能力 → 可追溯、可验证的循证医学支持
% [中文] 多模态能力：以文本医学理解为主 → 文本与图像融合的临床多模态感知
% [中文] 临床流程：以模型自身对话能力为主，尚未系统性纳入长期记忆与工具调用 → 动态问诊、随访、持续照护、工具调用与长期记忆协同
% [中文] 安全机制：基础安全对齐 → Harness 级动作约束、安全护栏与在线闭环迭代
\begin{table}[H]
\centering\small
\caption{Core capability upgrades from Baichuan-M3 to Baichuan-M4.}
\label{tab:capability-upgrades}
\begin{tabularx}{\linewidth}{lXX}
\toprule
\textbf{Dimension} & \textbf{Baichuan-M3} & \textbf{Baichuan-M4} \\
\midrule
Consultation capability & Multi-turn medical consultation focused on professional dialogue within individual care episodes & Dynamic consultation and follow-up for long-term, multi-stage continuous care \\
\addlinespace[2pt]
Evidence-based capability & Medical knowledge reasoning and experience-based judgment & Traceable and verifiable evidence-based medical support \\
\addlinespace[2pt]
Multimodal capability & Mainly text-based medical understanding & Integrated clinical multimodal perception across text and images \\
\addlinespace[2pt]
Clinical workflow & Primarily model-internal dialogue capabilities, without systematic integration of long-term memory or tool use & Dynamic consultation, follow-up visits, ongoing care, tool use, and long-term memory coordination \\
\addlinespace[2pt]
Safety mechanism & Basic safety alignment & Harness-level action constraints, safety guardrails, and online closed-loop iteration \\
\bottomrule
\end{tabularx}
\end{table}

% [中文] 1.3 预期应用
\subsection{Intended Applications}

% [中文] Baichuan-M4 面向严肃医疗场景设计，聚焦临床决策支持与智能医疗交互，广泛适用于以下核心任务：
\textbf{Baichuan-M4 is designed for serious medical scenarios, with a focus on clinical decision support and intelligent medical interaction. It is broadly applicable to the following core tasks:}

% [中文] 端到端照护支持：支持就诊前分诊、初诊、随访、慢病管理、用药反馈追踪与病情变化检测。
\paragraph{End-to-end care support:} Assists with pre-visit triage, initial consultation, follow-up visits, chronic disease management, medication feedback tracking, and detection of changes in disease progression.

% [中文] 临床流程与智能问答：辅助主诉分析、病史采集、患者风险分层、初步鉴别诊断、高危红旗症状识别，以及个性化就医建议生成。
\paragraph{Clinical workflow and intelligent Q\&A:} Assists with chief complaint analysis, medical history collection, patient risk stratification, preliminary differential diagnosis, identification of high-risk red-flag symptoms, and generation of personalized care-seeking recommendations.

% [中文] 循证医学决策支持：支持临床问题拆解、权威文献与指南检索、证据链总结，以及对可靠来源的可追溯引用。
\paragraph{Evidence-based medical decision support:} Supports clinical question decomposition, retrieval of authoritative literature and guidelines, evidence chain summarization, and traceable citation of reliable sources.

% [中文] 医疗文档结构化解析：准确识别并提取化验报告、处方、出院小结、影像报告等医疗文档中的关键临床信息。
\paragraph{Structured parsing of medical documents:} Accurately identifies and extracts key clinical information from laboratory reports, prescriptions, discharge summaries, imaging reports, and other medical documents.

% [中文] 多模态医学影像支持：理解 X 光、皮肤病变等多模态医学视觉信息，支持报告生成与诊断建议。
\paragraph{Multimodal medical imaging support:} Understands multimodal medical visual information such as X-rays and skin lesions, and supports report generation and diagnostic support suggestions.

% [中文] 使用范围：Baichuan-M4 定位为临床医生、医疗机构与合规医疗服务提供方的辅助工具。模型输出不得替代执业医师的独立专业判断，不应直接用于最终诊断、治疗方案制定、处方开具，或急危重症决策。
\textbf{Scope of use:} Baichuan-M4 is positioned as an \textbf{assistive tool for clinicians}, medical institutions, and compliant healthcare service providers. Model outputs must not replace the independent professional judgment of licensed physicians, and should not be used directly for final diagnosis, treatment planning, prescription issuance, or emergency and critical care decisions.

% [中文] 已知局限与风险：
\textbf{Known limitations and risks:}

\begin{enumerate}
% [中文] 长尾罕见病与非典型表现的局限：受医疗训练与评测数据分布影响，模型对部分罕见病、非典型症状、复杂共病或偏离常规临床路径的病例，诊断支持能力可能有限。
\item \textbf{Limitations in long-tail rare diseases and atypical presentations:} Due to the distribution of medical training and evaluation data, the model may have limited diagnostic support capability for rare diseases, atypical symptoms, complex comorbidities, or cases that deviate from common clinical pathways. Clinical review is required, especially when symptoms are persistent, progressive, or difficult to explain.

% [中文] 分诊与风险识别不完整的风险：虽然模型被设计用于识别红旗症状并提供就医建议，但仍可能遗漏、低估或高估部分急危重症风险。
\item \textbf{Risk of incomplete or incorrect triage:} Although Baichuan-M4 is designed to identify high-risk red-flag symptoms and provide care-seeking recommendations, it may still miss, under-prioritize, or over-prioritize certain urgent conditions. The model should not be used as the sole basis for emergency triage, critical care decisions, or decisions to delay medical attention.

% [中文] 药物与治疗建议风险：模型可能在禁忌证、药物相互作用、过敏史、剂量调整或特殊人群用药方面给出不完整或不适当的建议。
\item \textbf{Medication and treatment recommendation risks:} The model may generate incomplete or inappropriate suggestions related to medication use, contraindications, drug interactions, allergies, dosage adjustment, or special populations such as children, pregnant patients, older adults, and patients with liver or kidney impairment. Medication-related outputs must be reviewed by licensed clinicians and should not be used directly for prescription issuance or treatment modification.

% [中文] 长期记忆与患者状态追踪风险：连续照护依赖准确的患者记忆，但长期档案可能受到信息缺失、病史过期、用户自述不准确或记忆更新错误的影响。
\item \textbf{Risks in long-term memory and patient-state tracking:} Continuous care relies on accurate patient memory, but long-term profiles may be affected by incomplete information, outdated medical history, incorrect user self-reports, or errors during memory extraction and updating. Incorrect or stale memory may lead to inappropriate personalization. Important medical facts should be confirmed with the user or clinician before being used for clinical decision support.

% [中文] 循证检索与引用局限：外部检索可能返回过期、不完整、低质量或与具体患者情境不匹配的证据，引用也可能无法完全支撑生成结论。
\item \textbf{Evidence retrieval and citation limitations:} Evidence-based retrieval may still return outdated, incomplete, low-quality, or contextually mismatched sources. Even when citations are provided, the cited evidence may not fully support the generated conclusion, may not apply to the patient’s specific population or condition, or may conflict with newer guidelines. Clinicians should verify key evidence before using it in clinical decisions.

% [中文] 多模态感知与输入质量风险：在医学文档识别、X 光影像理解、皮肤病图像分析等任务中，分辨率、光照、角度、遮挡、设备差异、文档版式和人群偏差都可能影响输出质量。
\item \textbf{Multimodal perception and input-quality risks:} In document OCR, X-ray interpretation, dermatology image analysis, and other multimodal tasks, output quality may be affected by image resolution, lighting, angle, occlusion, device differences, document layout, scan quality, or population-level biases. Multimodal outputs should be treated as assistive signals rather than definitive diagnoses, and must be reviewed together with clinical context and professional examination.
\end{enumerate}

% [中文] 二、Baichuan-Harness：全新的医疗 Agent 系统
\section{Baichuan-Harness: A New Medical Agent System}

% [中文] 图1：Baichuan-Harness——训练与部署统一的运行时基座。
\begin{figure}[htbp]
    \centering
    \includegraphics[width=\linewidth]{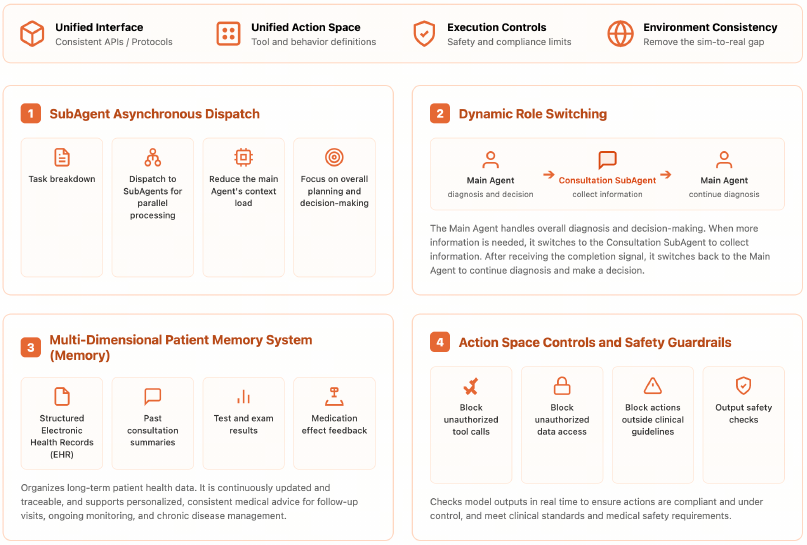}
    \caption{Baichuan-Harness, the unified runtime foundation for both training and deployment.}
    \label{fig:harness}
\end{figure}

% [中文] Baichuan-M4 不只是一个语言模型，更是一个高度协同的医疗 Agent 系统。如图1所示，系统以 Baichuan-Harness 作为训练与部署统一的运行时基座。在受控环境中，它支持模型行为约束、外部工具调用、长期患者记忆管理与多 Agent 协同。为缩小强化学习训练与真实部署之间的环境差异（即 Sim-to-Real Gap），M4 的强化学习训练完全在 Baichuan-Harness 内闭环完成。这确保了训练与部署在交互接口、动作空间与执行约束上的一致性。
Baichuan-M4 is not only a language model, but also a highly coordinated \textbf{medical agent system}~\cite{tao2024amie, elahe2025amie, liu2025medchainbridginggapllm}. As shown in Figure~\ref{fig:harness}, the system uses \textbf{Baichuan-Harness} as the unified runtime foundation for both training and deployment. In a controlled environment, \textbf{it enables model behavior constraints, external tool calls, long-term patient memory management, and multi-agent coordination.} To reduce the environment mismatch between reinforcement learning training and real-world deployment, also known as the Sim-to-Real Gap, the reinforcement learning training of M4 is completed entirely within Baichuan-Harness in a closed loop. \textbf{This ensures consistency between training and deployment in interaction interfaces, action space, and execution constraints.}

% [中文] Baichuan-Harness 负责以下核心功能与控制机制：
Baichuan-Harness is responsible for the following core functions and control mechanisms:

\begin{itemize}
    % [中文] 异步 SubAgent 调度：针对医学文献检索、证据链总结、长病史整理等高上下文负载任务，主 Agent 可将任务拆解并分派给专用 SubAgent 并行处理。该机制显著降低主 Agent 的上下文负担，使其能将算力集中于整体照护路径规划与决策生成。
    \item \textbf{Asynchronous SubAgent dispatch:} For high-context-load tasks such as medical literature retrieval, evidence chain summarization, and organization of long patient medical histories, the main Agent can decompose tasks and dispatch them to specialized SubAgents for parallel processing. This mechanism significantly reduces the context burden on the main Agent, allowing it to focus computational resources on overall care pathway planning and decision generation.
    % [中文] 动态角色切换：该机制用于支持高度标准化的临床交互流程。在多轮问诊中，负责诊断与决策的主 Agent 可暂时退至后台，由专门的问诊 SubAgent 高效收集信息；信息收集完成并触发结束信号后，系统切回主 Agent 进行综合分析。这种流式隔离机制有助于避免跨任务干扰，并提升在专业交互场景中的鲁棒性。
    \item \textbf{Dynamic role switching:} This mechanism is used to support highly standardized clinical interaction workflows. In multi-turn consultations, the main Agent, which is responsible for diagnosis and decision-making, can temporarily move to the background while a dedicated consultation SubAgent collects information efficiently. Once information collection is complete and an end signal is triggered, the system switches back to the main Agent for comprehensive analysis. This streaming isolation mechanism helps avoid cross-task interference and improves robustness in specialized interaction scenarios.
    % [中文] 多维患者记忆系统：连续照护依赖于对患者病情变化的长期理解。Harness 将患者病史组织为可追溯、动态更新的长期记忆，涵盖结构化电子病历、历史问诊摘要、既往化验与影像趋势，以及用药反馈。这使 M4 能在随访、持续照护与慢病管理中准确识别病情进展变化，并提供高度个性化、长期一致的医疗建议。
    \item \textbf{Multidimensional patient memory system:} Continuous care depends on long-term understanding of changes in a patient's condition. Harness organizes patient history as traceable and dynamically updated long-term memory, covering structured electronic health records, historical consultation summaries, trends from previous laboratory and imaging tests, and medication response feedback. This allows M4 to accurately identify changes in disease progression during follow-up visits, ongoing care, and chronic disease management, and to provide medical recommendations that are highly personalized and consistent over time.
    % [中文] 动作空间约束与安全护栏：作为医疗场景的安全合规层，Harness 实时校验模型的动作输出，严格限制非法工具调用、越权数据访问，以及不符合临床规范或照护路径的行为，从而确保系统在真实医疗环境中的可控性与安全性。
    \item \textbf{Action space constraints and safety guardrails:} As the safety and compliance layer for medical scenarios, Harness validates the model's action outputs in real time. It strictly restricts illegal tool calls, unauthorized data access, and behaviors that do not follow clinical standards or care pathways. This improves the controllability and safety of the system in controlled healthcare deployment environments.
\end{itemize}

% [中文] 三、模型：核心推理引擎
\section{Model: Core Reasoning Engine}

% [中文] 医疗大语言模型的可靠性无法仅靠静态监督微调（SFT）实现，它需要在连续照护场景中通过闭环反馈进行持续的自适应校准。
The reliability of a medical large language model cannot be achieved through static supervised fine-tuning (SFT) alone. It requires continuous adaptive calibration through closed-loop feedback in continuous care scenarios.

% [中文] M4 的模型设计超越了基于单一指标（如单轮 QA 准确率）的评估方式。相反，它将患者状态连续性、照护路径质量、医疗安全边界与真实临床反馈一并融入训练与推理系统。
The model design of M4 goes beyond evaluation based on a single metric, such as single-turn QA accuracy. Instead, \textbf{it integrates patient state continuity, care pathway quality, medical safety boundaries, and real clinical feedback into both the training and inference systems.}

% [中文] 图2：三层嵌套自适应控制系统，其持续改进由真实世界反馈驱动。
\begin{figure}[htbp]
    \centering
    \includegraphics[width=\linewidth]{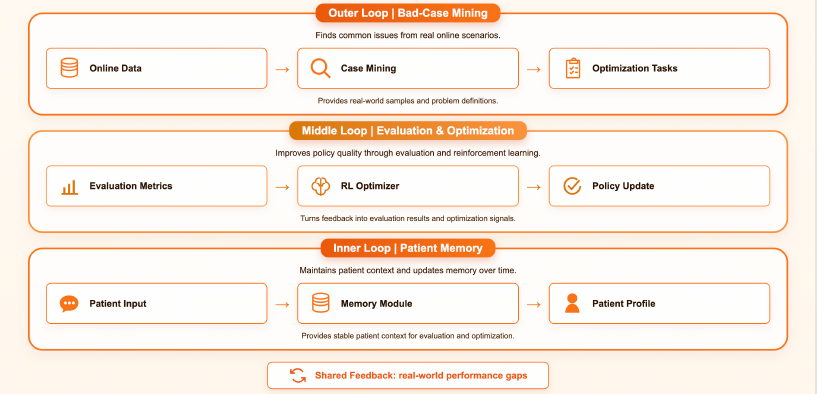}
    \caption{The three-layer nested adaptive control system, where continuous improvement is driven by real-world feedback.}
    \label{fig:closed-loop}
\end{figure}

% [中文] 3.1 面向连续照护的三层闭环
\subsection{Three-Layer Closed Loop for Continuous Care}

% [中文] 如图2所示，M4 围绕真实临床路径构建了三层自演进闭环架构：内环、中环与外环。这定义了系统持续改进的核心机制：
As illustrated in Figure~\ref{fig:closed-loop}, M4 builds a three-layer self-evolving closed-loop architecture around real clinical pathways: the inner loop, middle loop, and outer loop. This defines the core mechanism for continuous system improvement:

% [中文] 内环（患者级上下文连续性）：解决模型能否保留并正确衔接患者上下文的问题。模型能够将当前输入与患者既往病程整合。例如，当患者先报告"咳嗽与低热"，几天后又上传血常规并提供用药反馈时，模型能识别这是同一疾病发作的延续，进而结合历史信息与新证据进行综合评估，避免碎片化、孤立化的照护。
\paragraph{Inner loop (patient-level context continuity):} This addresses \textbf{whether the model can retain and correctly connect patient context}. The model can integrate the current input with the patient's prior disease course. For example, when a patient first reports ``cough and low-grade fever,'' then uploads a complete blood count and provides medication feedback several days later, the model can recognize that this is a continuation of the same disease episode. It can then combine historical information with new evidence for integrated assessment, avoiding fragmented and isolated care.

% [中文] 中环（临床级多维评估）：解决模型输出是否准确且受控的问题。评估范围从单轮生成质量扩展到完整的连续照护路径。通过内部自动化评估与医学专家评审相结合，系统对医疗准确性、逻辑一致性、安全护栏合规性、主动澄清行为等多个维度进行细致衡量，并将这些评估指标直接转化为强化学习的对齐信号。

\paragraph{Middle loop (clinical-level multidimensional evaluation):} This addresses \textbf{whether the model's output is accurate and well controlled}. The evaluation scope is expanded from single-turn generation quality to the full continuous care pathway. Through both in-house automated evaluation and medical expert review, the system measures medical accuracy, logical consistency, compliance with safety guardrails, active clarification behavior, and other dimensions in detail. These evaluation metrics are then converted directly into alignment signals for reinforcement learning.

% [中文] 外环（真实场景中的在线自演进）：解决系统如何主动发现长尾问题并持续改进的问题。外环捕捉在线 bad case、用户追问、医生纠正、低置信回答与工具调用异常等弱信号；经过自动脱敏、语义聚类、根因分析与专家校验后，这些信号被转化为高价值评估样本与训练数据，驱动模型安全、快速的在线迭代。
\paragraph{Outer loop (online self-evolution in real-world scenarios):} This addresses \textbf{how the system actively identifies long-tail issues and keeps improving}. The outer loop captures weak signals such as online bad cases, user follow-up questions, physician corrections, low-confidence responses, and tool-use exceptions. After automated de-identification, semantic clustering, root-cause analysis, and expert validation, these signals are converted into high-value evaluation samples and training data, driving safe and fast online iteration of the model.

% [中文] 3.2 强化学习与训练算法创新
\subsection{Reinforcement Learning and Training Algorithm Innovations}

% [中文] M4 的能力提升源自一个面向连续照护场景设计的通用强化学习框架。该框架将问诊、循证推理、多模态感知与工具规划统一到同一套策略优化体系中，使模型在医疗严谨性、交互效率与安全边界之间达到更优的帕累托前沿。
The capability improvement of M4 comes from a \textbf{general reinforcement learning framework designed for continuous care scenarios}. This framework unifies consultation, evidence-based reasoning, multimodal perception, and tool planning under the same policy optimization system. It enables the model to reach a better Pareto frontier across medical rigor, interaction efficiency, and safety boundaries.

% [中文] SPAR++：质量优先的片段级奖励建模。M4 将基于评分量表（rubric）的强化学习范式升级为 SPAR++ 算法。为解决复杂 Agent 任务中的信用分配（credit assignment）问题，SPAR++ 用锚定在关键临床片段上的奖励信号，取代对整段对话轨迹的粗粒度打分。模型不仅因得出正确最终结论而获奖励，也因充分的病史采集、及时的风险识别与恰当的工具使用而获奖励。SPAR++ 还引入质量门控（quality gating）机制：诸如减少问诊轮次之类的效率奖励，只有在医疗质量分达到所需阈值时才生效，从而避免模型为更快结束问诊而省略关键病史。
\paragraph{SPAR++: quality-first span-level reward modeling.} M4 upgrades the rubric-based reinforcement learning paradigm~\cite{baichuan-m2, DBLP:journals/corr/abs-2508-12790, DBLP:journals/corr/abs-2507-17746} into the SPAR++ algorithm. To address the \textbf{credit assignment} problem in complex agent tasks, SPAR++ replaces coarse-grained scoring of an entire dialogue trajectory with reward signals anchored to key clinical spans. The model is not only rewarded for reaching the correct final conclusion, but also for sufficient history taking, timely risk identification, and appropriate tool use. SPAR++ also introduces a \textbf{quality gating} mechanism: efficiency rewards, such as reducing the number of consultation turns, take effect only when the medical quality score meets the required threshold. This prevents the model from omitting key medical history simply to close the consultation faster.
% [中文] 推理路径压缩。为降低在线推理时延（TTFT）并节省上下文空间，M4 引入推理链压缩机制。在不损害模型推理准确性的前提下，将每次内部推理过程（reasoning_content）的 token 消耗降至原来的六分之一。该技术利用强化学习对冗余推理施加细粒度惩罚，仅约束内部思维链的长度，而不压缩面向用户的最终回复（content）。由此释放出宝贵的上下文窗口容量，用于支撑更长期的患者记忆与循证检索内容。

% TODO-CITE: 可选，为"推理路径压缩"补高效推理/思维链压缩参考（bib 暂无，建议 key: efficient_reasoning）
\paragraph{Reasoning Path Compression.} To reduce online inference latency (TTFT) and save context space, M4 introduces a reasoning-chain compression mechanism. Without degrading the model's reasoning accuracy, it reduces the token consumption of each internal reasoning process (reasoning\_content) to \textbf{one-sixth of the original level.} This technique uses reinforcement learning to apply fine-grained penalties to redundant reasoning, constraining only the length of the internal chain of thought while leaving the final user-facing response (content) uncompressed. As a result, valuable context-window capacity is freed up to support longer-term patient memory and evidence-based retrieval content.
% [中文] 面向连续照护的课程式强化学习（Curriculum RL）。在初诊与随访混合的场景中，M4 采用"先以初诊打基础、再以随访提性能"的课程学习策略。第一阶段巩固基础症状识别与标准化病史采集能力；第二阶段混入多源异构的 Memory 数据（如结构化电子病历与非结构化对话摘要），以增强模型在不完整信息下对疾病进展进行推理并动态调整策略的能力。

\paragraph{Curriculum RL for Continuum of Care.} In mixed initial-visit and follow-up scenarios, M4 uses a curriculum learning strategy~\cite{bengio2009curriculum} of ``building the foundation with initial visits first, then improving performance with follow-ups.'' The first stage consolidates basic symptom recognition and standardized history-taking capabilities. The second stage mixes multi-source, heterogeneous Memory data (such as structured EHRs and unstructured dialogue summaries) to strengthen the model's ability to reason about disease progression and adjust strategies dynamically under incomplete information.

% [中文] SAPO 与 R3 路由回放机制。为应对大规模强化学习中的训练不稳定，M4 引入 SAPO 算法，以更平滑的策略梯度更新取代硬裁剪，有效降低策略震荡；并结合 R3 路由回放机制，抑制训练中的损失尖峰（Loss Peak）并施加 KL 散度约束，防止熵坍缩，确保在高并发、多轮交互场景下策略训练的稳定性。
\paragraph{SAPO and R3 Route Replay Mechanism.} To address training instability in large-scale reinforcement learning, M4 introduces the \textbf{SAPO algorithm}~\cite{DBLP:journals/corr/abs-2509-02333, DBLP:journals/corr/abs-2507-18071, guo2025deepseek}, which replaces hard clipping with a smoother policy-gradient update and effectively reduces policy oscillation. It is also combined with the \textbf{R3 route replay mechanism} to suppress loss peaks (Loss Peak) during training and apply KL-divergence constraints, preventing entropy collapse and ensuring stable policy training in high-concurrency, multi-turn interaction scenarios.

% [中文] 四、工具：临床级感知、知识检索与状态管理系统
\section{Tool: Clinical-Grade Perception, Knowledge Retrieval, and State Management System}

% [中文] 工具层是 M4 与真实物理世界交互的"感官"与"双手"。它使 M4 能够超越纯参数化记忆，在临床实践中访问长期患者记录、解析异构医学影像，并检索权威的循证参考。
The Tool layer serves as M4's ``senses'' and ``hands'' for interacting with the real physical world. It allows M4 to go beyond purely parametric memory by accessing long-term patient records, parsing heterogeneous medical images, and retrieving authoritative evidence-based references in clinical practice.

% [中文] 4.1 患者记录与动态记忆管理工具
\subsection{Patient Records and Dynamic Memory Management Tools}

% [中文] 为确保个性化、一致的连续照护，M4 建立了严格的记忆管理机制，将瞬时会话上下文与长期患者档案解耦：
To ensure personalized and consistent continuous care, M4 establishes a rigorous memory management mechanism that decouples transient session context from the long-term patient profile:

% [中文] 短期记忆（会话上下文）：聚焦当前会话内的上下文流，包括瞬时症状、患者当前主诉、临时会话目标与尚未确认的生理细节，用于在多轮交互中保持自然连贯。
\paragraph{Short-term Memory (Session Context):} Focuses on the context flow within the current session, including transient symptoms, the patient's current chief complaint, temporary session goals, and unconfirmed physiological details. It is used to maintain natural coherence across multi-turn interactions.

% [中文] 长期记忆（长期档案）：诸如慢病诊断、手术史、严重过敏反应与关键化验趋势等已确认的医学事实，需经过"抽取、置信度评估、必要时用户确认"的合规链路后，才写入长期结构化档案（Profile）。模糊描述或未经核实的自述不得污染长期记忆。
\paragraph{Long-term Memory (Long-term Profile):} Confirmed medical facts, such as chronic disease diagnoses, surgical history, severe allergic reactions, and key laboratory trends, are written into the long-term structured Profile only after passing a compliant chain of \textbf{``extraction, confidence assessment, and user confirmation when needed.''} Vague descriptions or unverified self-reports are not allowed to contaminate long-term memory.

% [中文] 渐进式隐私与上下文披露机制（Progressive Disclosure）：为保护患者隐私并减少模型推理中无关噪声的干扰，系统默认仅向 Agent 暴露长期档案摘要与当前最相关的短期上下文。只有当 Agent 发出深度回溯指令时，工具链才按需加载底层原始电子病历或历史报告。这在"最小隐私暴露"与"最大计算效率"之间取得平衡。
\paragraph{Progressive Privacy and Context Disclosure Mechanism (Progressive Disclosure):} To protect patient privacy and reduce interference from irrelevant noise during model reasoning, the system by default exposes only the long-term Profile summary and the short-term context most relevant to the current situation to the Agent. Only when the Agent issues a deep backtracking command does the toolchain load the underlying raw electronic medical records or historical reports as needed. This balances ``minimum privacy exposure'' with ``maximum computational efficiency.''

% [中文] 4.2 权威循证医学检索工具
\subsection{Authoritative Evidence-Based Medical Retrieval Tool}

% [中文] 循证医学要求每条陈述都可追溯到可靠来源。在数据源层面，M4 过滤掉非专业公开网页，构建了结构严密的六层权威医学证据金字塔：
Evidence-based medicine requires every statement to be traceable to a reliable source. At the data-source level, M4 filters out non-professional public webpages and builds a tightly structured \textbf{six-layer pyramid of authoritative medical evidence:}

\begin{itemize}
    % [中文] 一级研究层：索引海量医学期刊论文，收录超过 4000 万篇文献，规模超过 PubMed 的索引量；覆盖基础与临床研究成果，作为证据链的起点。
    \item \textbf{Primary Research Layer:} Indexes a large volume of medical journal articles, with more than 40 million biomedical records, including journal articles and related scholarly metadata. It covers both basic and clinical research findings and serves as the starting point of the evidence chain.
    % [中文] 证据综述层：整合系统综述、meta 分析等高级别证据，提供经过整合的研究结论。
    \item \textbf{Evidence Review Layer:} Integrates high-level evidence such as systematic reviews and meta-analyses, providing consolidated research conclusions.
    % [中文] 指南与标准层：纳入由国内外权威机构发布的临床指南、专家共识与行业标准，确保回答遵循最新临床规范。
    \item \textbf{Guideline and Standard Layer:} Incorporates clinical guidelines, expert consensus, and industry standards issued by authoritative international and domestic organizations, ensuring that responses follow the latest clinical norms.
    % [中文] 实践知识层：包含临床病例报告、一线专家经验与诊疗技巧等实践知识，更贴近真实医疗实践。
    \item \textbf{Practical Knowledge Layer:} Includes practical knowledge such as clinical case reports, frontline expert experience, and diagnostic and treatment techniques, making it closer to real-world medical practice.
    % [中文] 公共健康教育层：汇聚权威的健康教育与公共卫生知识，如疾病预防资料与健康指导，以支持公众健康教育。
    \item \textbf{Public Health Education Layer:} Aggregates authoritative health education and public health knowledge, such as disease prevention materials and health guidance, to support public health education.
    % [中文] 监管与真实世界证据层：涵盖药品监管公告、临床试验注册与大规模真实世界研究数据，反映最新监管动态与人群层面的研究发现。
    \item \textbf{Regulatory and Real-World Evidence Layer:} Covers drug regulatory announcements, clinical trial registries, and large-scale real-world study data, reflecting the latest regulatory updates and population-level research findings.
\end{itemize}

% [中文] 为在庞大而复杂的知识库中实现精准检索，M4 引入 PICO（人群、干预、对照、结局）拆解策略，并利用强化学习对 Query 生成策略进行对齐。系统将相关性、时效性、权威性与检索结果的 PICO 契合度等多维奖励信号注入验证器系统（Verifier System），使 M4 在检索召回精度与证据链组装能力上远超通用大模型的 RAG 表现。
To achieve precise retrieval from a large and complex knowledge base, M4 introduces a \textbf{PICO (Population, Intervention, Comparison, Outcome) decomposition strategy}~\cite{schardt2007utilization} and uses reinforcement learning to align the Query generation strategy. The system injects multi-dimensional reward signals, including relevance, timeliness, authority, and PICO fit of retrieval results, into the \textbf{Verifier System}~\cite{baichuan-m2}. This enables M4 to achieve much higher retrieval recall precision and evidence-chain assembly capability than the RAG performance of general-purpose LLMs~\cite{lewis2020retrieval}.

% [中文] 4.3 多模态医学感知工具
\subsection{Multimodal Medical Perception Tools}

% [中文] M4 在工具层深度集成视觉语言模型（VLM）的感知能力，将医疗文档、放射影像、病变图像等异构临床视觉信号转化为可靠的结构化医学表示。
M4 deeply integrates vision-language model (VLM) perception capabilities~\cite{MedGemma, Lingshu, Hulu-Med} at the Tool layer. It converts heterogeneous clinical visual signals, including medical documents, radiological images, and lesion images, into reliable structured medical representations.

\begin{itemize}
    % [中文] 高保真医疗文档识别（OCR）：针对血常规报告、生化检验报告、处方等高度异构的表格数据，M4 不采用强制模型产出预定义版式的传统做法，而是使用基于非结构化相似度与结构化二部匹配的双重评估框架。由一个轻量辅助模型将 VLM 输出转换为标准化格式，用于核心键值对匹配。这避免了超大表格中因格式 token 溢出导致的解码失败，确保对异常箭头、检验数值、单位等关键临床指标的高保真提取。
    \item \textbf{High-fidelity medical document recognition (OCR):} For highly heterogeneous tabular data such as complete blood count reports, biochemical test reports, and prescriptions, M4 avoids the traditional approach of forcing the model to produce a predefined layout. Instead, \textbf{it uses a two-part evaluation framework based on unstructured similarity and structured bipartite matching.} An LLM converts VLM outputs into a standardized format for core key-value pair matching. This prevents decoding failures caused by format token overflow in very large tables and ensures high-fidelity extraction of key clinical indicators, such as abnormality arrows, test values, and units.
    
    \item \textbf{X-ray report generation.} Supports clinical-grade assistive analysis of chest X-ray images by generating structured radiology reports, including both \textbf{Findings} (observations) and \textbf{Impression} (diagnostic opinion). Results on the IU-Xray benchmark demonstrate the model's capability in medical image understanding and report generation.

    % [中文] 面向皮肤病的临床证据驱动 Agent：M4 克服了传统 VLM 依赖"单图、闭卷分类"的局限，将皮肤病分析建模为证据驱动的多步决策过程。系统构建了涵盖候选生成、假设验证、视觉类比、形态学概念与鉴别诊断的五维医学证据空间，并利用强化学习优化 Agent 的证据获取路径。系统还引入工具调用必要性约束：缺乏完整证据链的诊断不获奖励，并对证据误用施加惩罚。这使模型能够规划出类似临床专家的循证决策路径：先检查、再对比、最后鉴别，从而使整个诊断过程可追溯、可审计。
    \item \textbf{Clinical evidence-driven agent for dermatology:} M4 overcomes the limitations of traditional VLMs that rely on ``single-image, closed-book classification.'' \textbf{It models dermatological analysis as an evidence-driven, multi-step decision process.} The system builds a five-dimensional medical evidence space covering candidate generation, hypothesis verification, visual analogy, morphological concepts, and differential diagnosis. Reinforcement learning is used to optimize the agent's evidence acquisition path. The system also \textbf{introduces a tool-calling necessity constraint}, where diagnoses without a complete evidence chain receive no reward, along with penalties for evidence misuse. This allows the model to plan an evidence-based decision path similar to that of a clinical expert: first examine, then compare, and finally differentiate. This makes the full diagnostic process traceable and auditable.
\end{itemize}

% [中文] 五、评测框架与定量结果
\section{Evaluation Framework and Quantitative Results}

% [中文] 为全面评估 Baichuan-M4 在严肃医疗部署场景中的核心能力，我们构建了一套跨维度的多模态医学评测框架。评测涵盖四大核心模块：语言与临床流程能力、面向循证医学的工具调用、医疗文档 OCR，以及多模态医学影像理解。
To comprehensively evaluate \textbf{Baichuan-M4}'s core capabilities in serious medical deployment scenarios, we developed a cross-dimensional multimodal medical evaluation framework. The evaluation covers four core modules: \textbf{language and clinical workflow capability, tool use for evidence-based medicine, medical document OCR, and multimodal medical image understanding.}

% [中文] 5.1 评测基准概览
\subsection{Benchmark Overview}

% [中文] 模型在下表所汇总的、经学术界或临床专家验证的权威基准上进行了广泛测试。
The model was extensively tested on the authoritative benchmarks summarized in Table~\ref{tab:benchmark-overview}, validated by the academic community or clinical experts.

% [中文] 表3：评测基准概览。
% [中文] 表头：能力模块 | 基准 | 定位与评测重点
% [中文] 语言与临床流程 —— HealthBench：评估静态医学知识、复杂高难度临床决策与基础多轮问答。
% [中文] Scan-Bench V1：动态全流程初诊场景；模拟标准 OSCE 临床考试，评估病史采集与初步诊断。
% [中文] Scan-Bench V2：病情持续进展的动态随访场景；评估对临床病程的时序理解、治疗反馈与基于风险的安全分诊。
% [中文] 工具调用能力 —— Baichuan-EBM：循证医学检索；基于 657 道专家标注题，评估模型调用外部引擎并检索密集临床证据的能力。
% [中文] Baichuan-Med-OCR：医疗文档数字化；评估对化验报告、出院小结、影像报告等复杂医疗文本的结构化抽取。
% [中文] X-Ray / Pathology：在 ChestDR、MIMIC-CXR、ROCOv2 等七大公开数据集上的宏平均，用于评估医学影像报告生成。
% [中文] f17k Benchmark：包含 4893 例真实皮肤病病例，
\begin{table}[H]
\centering\small
\caption{Overview of the evaluation benchmarks.}
\label{tab:benchmark-overview}
\begin{tabularx}{\linewidth}{@{}>{\raggedright\arraybackslash}p{2.4cm} l X@{}}
\toprule
\textbf{Capability Module} & \textbf{Benchmark} & \textbf{Positioning and Evaluation Focus} \\
\midrule
\multirow{3}{=}{Language and Clinical Workflow}
 & HealthBench & Evaluates static medical knowledge, complex high-difficulty clinical decision-making, and basic multi-turn question answering. \\
\cmidrule(l){2-3}
 & Scan-Bench V1 & Dynamic full-process initial consultation scenario. It simulates a standard OSCE clinical examination and evaluates history taking and preliminary diagnosis. \\
\cmidrule(l){2-3}
 & Scan-Bench V2 & Dynamic follow-up scenario with continuous disease progression. It evaluates temporal understanding of the clinical course, treatment feedback, and risk-based safety triage. \\
\midrule
\multirow{4}{=}{Tool Use Capability}
 & Baichuan-EBM & Evidence-based medical retrieval. Based on 657 expert-annotated questions, it evaluates the model's ability to call external engines and retrieve dense clinical evidence. \\
\cmidrule(l){2-3}
& Baichuan-Med-OCR & A benchmark for evaluating medical document understanding and information extraction from clinical texts, including lab reports, discharge summaries, and radiology reports. The model outputs are post-processed into structured representations using an LLM for evaluation. \\
\cmidrule(l){2-3}
& IU-Xray & A standard benchmark for chest X-ray report generation, used to evaluate medical image understanding and report generation capabilities. \\
\cmidrule(l){2-3}
& f17k Benchmark & A curated subset of the f17k dataset consisting of 4{,}893 labeled dermatology cases, obtained by randomly sampling 30\% of the original dataset and removing unlabeled images. It is used to evaluate multimodal agent performance in disease diagnosis. \\
\bottomrule
\end{tabularx}
\end{table}

% [中文] 5.2 医学语言与临床流程
\subsection{Medical Language \& Clinical Workflow}

% [中文] 本模块从两个子维度分析：静态医学基础能力（含安全关键指标），以及动态全流程临床问诊能力。
This module is analyzed across two sub-dimensions: static medical foundation capabilities, including safety-critical metrics, and dynamic full-process clinical consultation capability.

% [中文] 5.2.1 静态医学知识与安全指标
\subsubsection{Static Medical Knowledge \& Safety Metrics}

\begin{figure}[H]
\centering
\includegraphics[width=\linewidth]{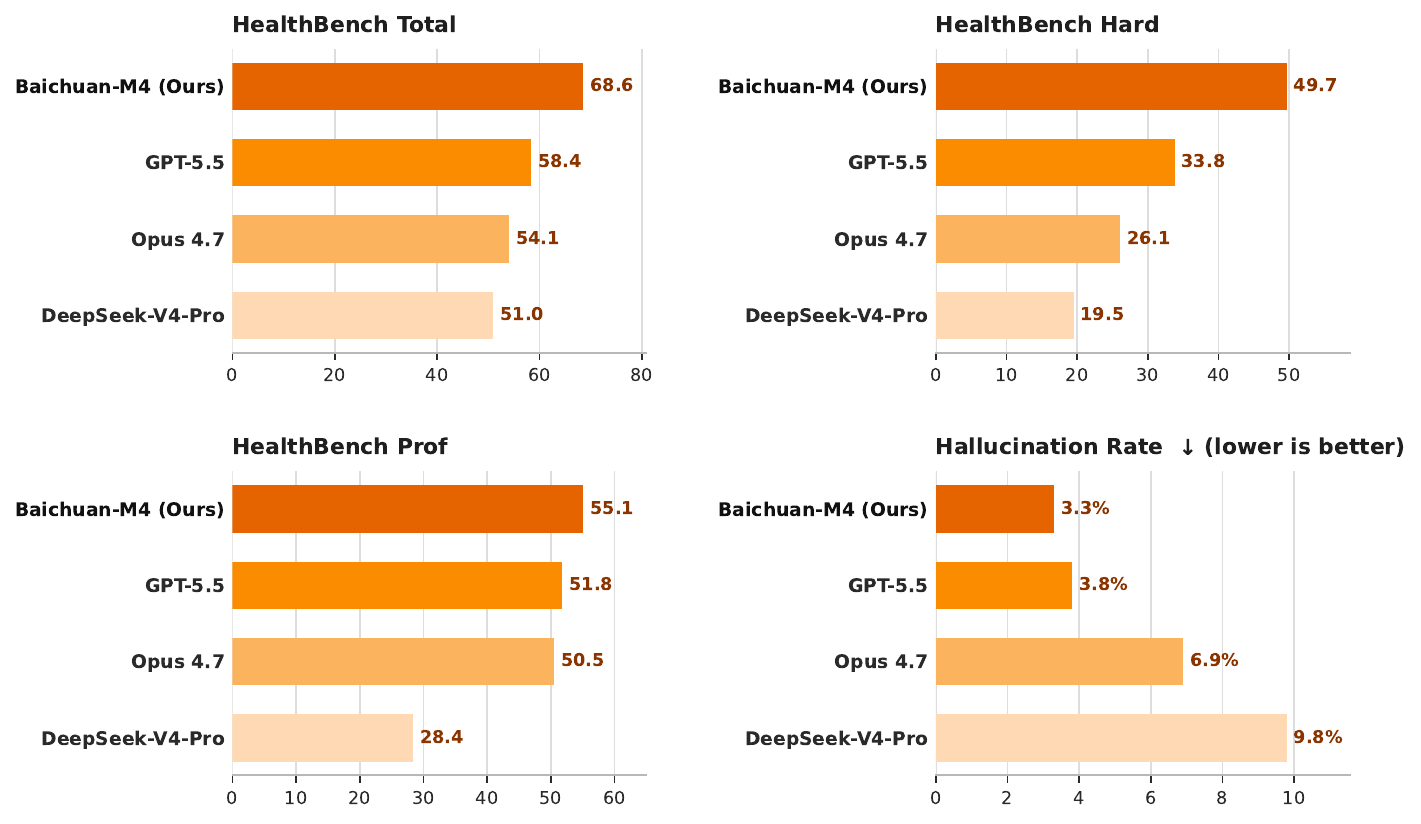}
\caption{Static medical knowledge and safety metrics across the four models, shown in Baichuan's brand palette. Higher is better for the three HealthBench panels, while lower is better for the hallucination rate; Baichuan-M4 leads every knowledge metric and attains the lowest hallucination rate (3.3\%).}
\label{fig:static-safety}
\end{figure}

% [中文] 本节聚焦模型的静态医学知识、高难度临床推理能力，以及严肃医疗部署所需的关键安全基线：幻觉率。（脚注：除幻觉率外，所有指标均为分数，越高越好。）
This section focuses on the model's static medical knowledge, high-difficulty clinical reasoning ability, and the key safety baseline required for serious medical deployment: hallucination rate~\cite{DBLP:journals/corr/abs-2503-05777, ankit2023medhalt}.\footnote{Except for Hallucination Rate, all metrics are scores, where higher is better.} As shown in Figure~\ref{fig:static-safety}, we compare these metrics across the evaluated models.

% [中文] 图：表4 静态医学知识与安全指标的可视化（柱状图，采用公司橙色品牌配色）。
% [中文] 三个 HealthBench 面板越高越好，幻觉率面板越低越好；Baichuan-M4 在三项知识指标上均领先，且幻觉率最低（3.3%）。
% [中文] 图源脚本：make_table4_chart.py（数据来自原 Table 4）。

% [中文] 原始表格代码（已转为上方图片，保留以便回退）：
% \begin{table}[H]
% \centering\small
% \caption{Static medical knowledge and safety metrics.}
% \label{tab:static-safety}
% \begin{tabularx}{\linewidth}{@{}l CCCC@{}}
% \toprule
% \textbf{Model} & \textbf{HealthBench Total} & \textbf{HealthBench Hard} & \textbf{HealthBench Prof} & \textbf{Hallucination Rate $\downarrow$} \\
% \midrule
% \textbf{Baichuan-M4 (Ours)} & \textbf{68.6} & \textbf{49.7} & \textbf{55.1} & \textbf{3.30\%} \\
% GPT-5.5 & 58.4 & 33.8 & 51.8 & 3.80\% \\
% Opus 4.7 & 54.1 & 26.1 & 50.5 & 6.90\% \\
% DeepSeek-V4-Pro & 51 & 19.5 & 28.4 & 9.80\% \\
% \bottomrule
% \end{tabularx}
% \end{table}

\begin{itemize}
    % [中文] 复杂决策的明显领先：在评估高风险与急症场景复杂推理的 HealthBench Hard 子集上，Baichuan-M4 领先次优模型 GPT-5.5 达 15.9 分，体现出更强的危机处理与医学逻辑推理能力。
    \item \textbf{Clear lead in complex decision-making:} On the HealthBench Hard subset~\cite{healthbench}, which evaluates complex reasoning in high-risk and emergency scenarios, Baichuan-M4 leads the second-best model, GPT-5.5~\cite{openai2026gpt55}, by \textbf{15.9} points. This demonstrates stronger crisis handling and medical logical reasoning capabilities.
    % [中文] 对安全关键风险的严格控制：M4 将严肃医疗应用中最关键的风险之一——幻觉率降至 3.3%，显著优于开源模型与通用闭源模型，大幅提升真实临床部署的安全裕度。
    \item \textbf{Strict control of safety-critical risks:} M4 reduces the hallucination rate, one of the most critical risks in serious medical applications, \textbf{to 3.3\%.} This is significantly better than both open-source models and general-purpose closed-source models, improving the safety margin for real-world clinical deployment.
\end{itemize}

% [中文] 5.2.2 动态临床问诊与记忆容量评测
\subsubsection{Dynamic Clinical Consultation \& Memory Capacity Evaluation}

\begin{figure}[H]
\centering
\includegraphics[width=\linewidth]{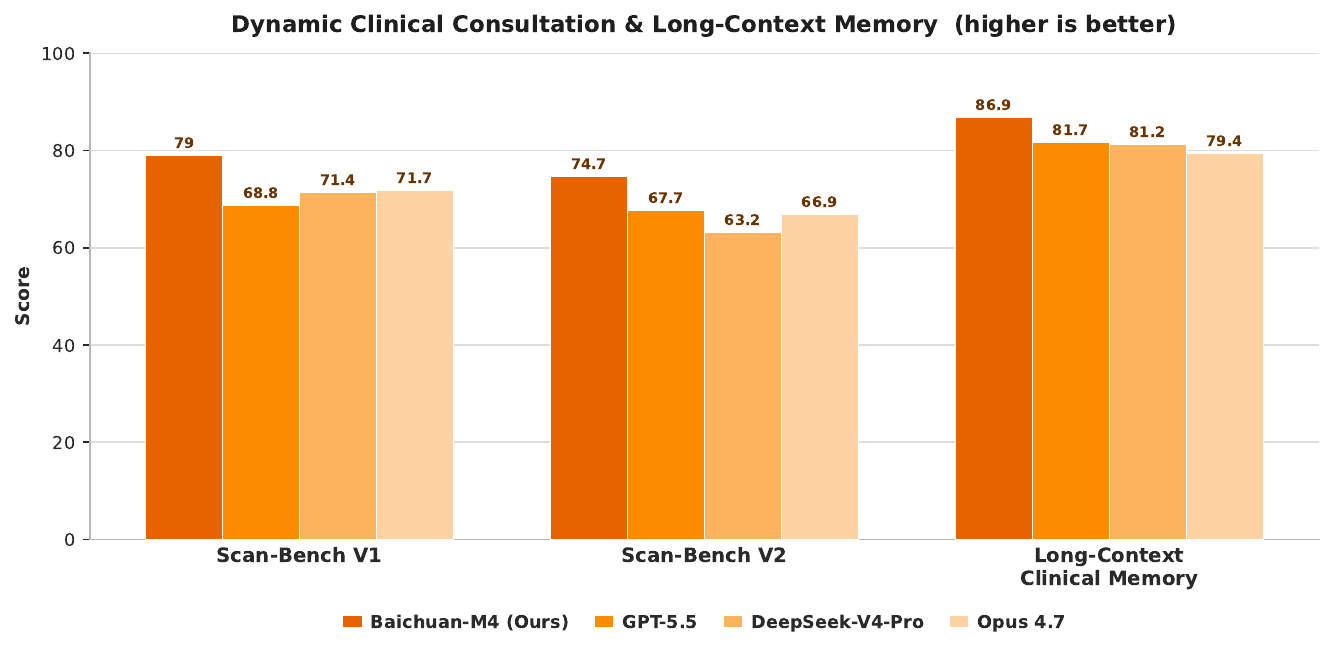}
\caption{Dynamic clinical consultation and long-context memory capacity across models (higher is better). Baichuan-M4 leads on both Scan-Bench V1 and V2 and shows its widest margin on long-context clinical memory, reaching 86.9 versus 79.4--81.7 for the other models.}
\label{fig:dynamic-memory}
\end{figure}

% [中文] 本节展示模型在模拟真实 OSCE 临床考试中的表现，涵盖初诊与随访，并引入针对长临床对话的专门"记忆容量"评测。
This section presents the model's performance in simulated real-world OSCE clinical examinations, covering both initial visits and follow-up visits. It also introduces a dedicated ``memory capacity'' evaluation for long clinical conversations. Figure~\ref{fig:dynamic-memory} reports these results across models.

% [中文] 图：表4（动态临床问诊与长上下文记忆）数据的可视化——分组柱状图（与图 3 不同的形式），公司橙色配色。
% [中文] Scan-Bench V1/V2 与长上下文临床记忆三项指标，越高越好；Baichuan-M4 三项均领先，长上下文记忆优势最明显（86.9）。
% [中文] 图源脚本：make_dynamic_memory_chart.py（数据来自原动态记忆表）。

% [中文] 原始表格代码（已转为上方图片，保留以便回退）：
% \begin{table}[H]
% \centering\small
% \caption{Dynamic clinical consultation and long-context memory capacity (higher is better).}
% \label{tab:dynamic-memory}
% \begin{tabularx}{\linewidth}{@{}l CCC@{}}
% \toprule
% \textbf{Model} & \textbf{Scan-Bench V1} & \textbf{Scan-Bench V2} & \textbf{Long-Context Clinical Memory} \\
% \midrule
% \textbf{Baichuan-M4} & \textbf{79} & \textbf{74.7} & \textbf{86.9} \\
% GPT-5.5 & 68.8 & 67.7 & 81.7 \\
% DeepSeek-V4-Pro & 71.4 & 63.2 & 81.2 \\
% Opus 4.7 & 71.7 & 66.9 & 79.4 \\
% \bottomrule
% \end{tabularx}
% \end{table}

\begin{itemize}
    % [中文] 动态问诊的持续领先：在需要模型主动提问并逐步进行鉴别筛查的 Scan-Bench V1/V2 场景中，Baichuan-M4 在医疗流程执行上持续展现出明显的原生优势，弥补了通用模型常陷入被动应答模式的普遍缺陷。
    \item \textbf{Sustained leadership in dynamic consultation:} In \textbf{Scan-Bench V1/V2} scenarios, where the model must actively ask questions and perform step-by-step differential screening, Baichuan-M4 consistently shows a clear native advantage in medical workflow execution. It addresses the common limitation of general-purpose models, which often fall into passive response patterns.
    % [中文] 新增：长上下文临床记忆的重大突破（+21.1）：作为本代核心升级，M4 在长上下文记忆维度取得 86.9 的最高分，较第一代 Baichuan-M3 的 65.8 大幅提升 21.1 分，并超过 GPT-5.5（81.7）与 DeepSeek-V4-Pro（81.2）。这一提升确保模型在审阅大量历史病历或进行长程多轮临床问诊等复杂场景中，不会遗漏先前提及的任何关键"红旗症状"或相关病史。
    \item \textbf{New: major breakthrough in long-context clinical memory (+21.1):} As a core upgrade of this generation, M4 achieves the \textbf{highest score of 86.9} in the long-context memory dimension. This is a substantial 21.1-point improvement over the first-generation Baichuan-M3 score of 65.8, and it also surpasses GPT-5.5 (81.7) and DeepSeek-V4-Pro~\cite{deepseekai2026deepseekv4} (81.2). This improvement ensures that the model does not miss any key ``red flag symptoms'' or relevant medical history mentioned earlier when handling complex scenarios such as reviewing extensive historical medical records or conducting long multi-turn clinical consultations.
\end{itemize}

% [中文] 5.3 工具调用与多模态能力评测
\subsection{Tool Calling \& Multimodal Capability Evaluation}

% [中文] 5.3.1 循证医学
\subsubsection{Evidence-Based Medicine}

% [中文] 本评测衡量模型通过外部搜索引擎调用所获取的医学证据质量。针对每个查询，专家标注关键检索目标，称为 Nuggets。
This evaluation measures the quality of medical evidence obtained by the model through external search engine calls. For each query, experts annotate the key retrieval targets, referred to as \textbf{Nuggets}~\cite{voorhees2003overview}.

\begin{itemize}
    % [中文] 核心分（Core Score）：衡量检索到的外部证据能否真正支撑核心医学 Nuggets。
    \item \textbf{Core Score:} Measures whether the retrieved external evidence can truly support the core medical Nuggets.
    % [中文] 引用精确率（Citation Precision）：衡量引用准确性与证据冗余度；分数越高，表示信息密度越高、无关或填充式证据越少。
    \item \textbf{Citation Precision:} Measures citation accuracy and evidence redundancy. A higher score indicates higher information density and less irrelevant or padded evidence.
\end{itemize}

\begin{figure}[H]
\centering
\includegraphics[width=0.85\linewidth]{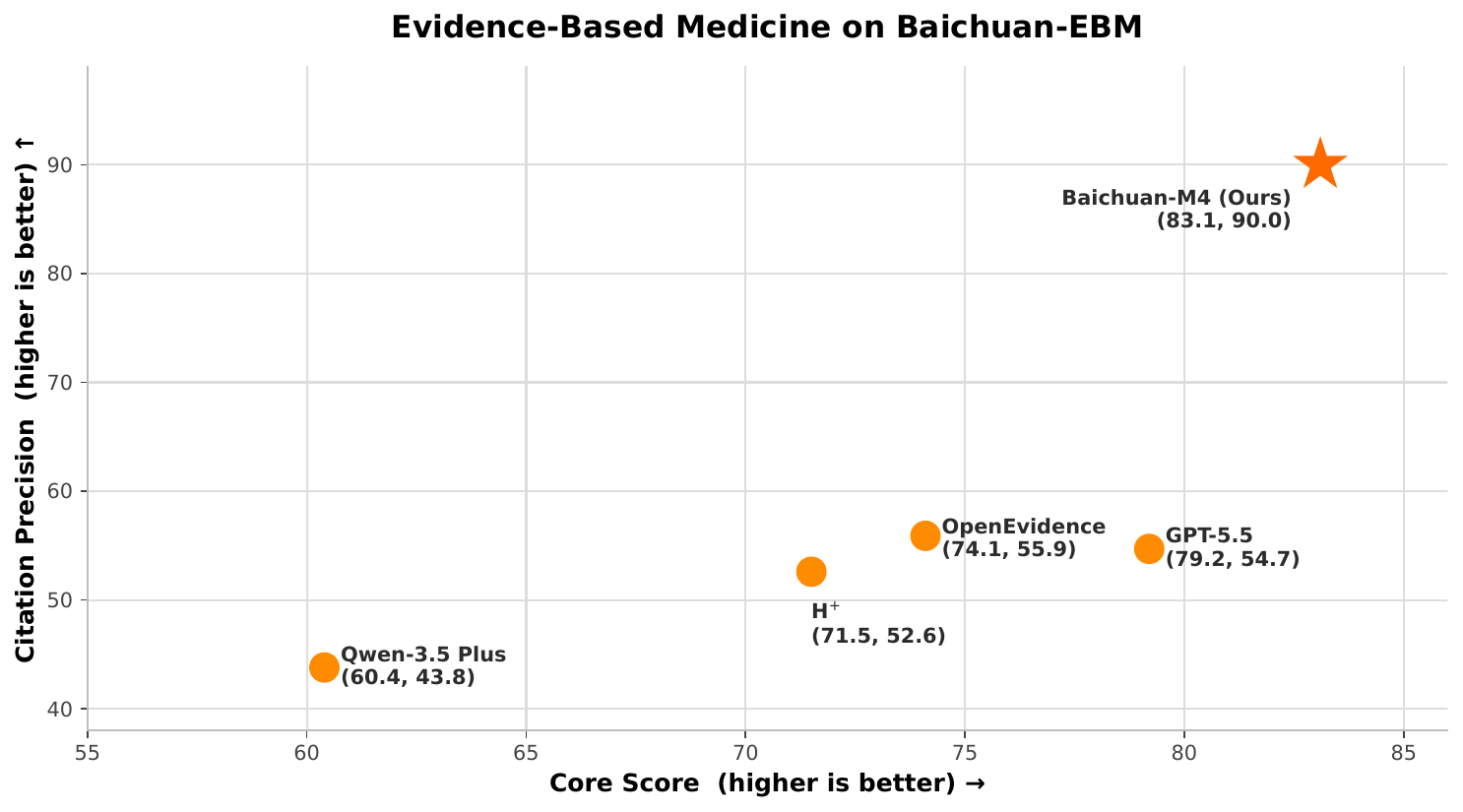}
\caption{Evidence-based medicine results on Baichuan-EBM, plotting Core Score against Citation Precision (higher is better on both axes). $H^+$ stands for AI-Doctor of Alibaba Group. Website: \url{https://www.ali-doctor.com/}. Baichuan-M4 occupies the top-right corner, leading on both axes, with an especially large lead in citation precision (90 vs.\ 43.8--55.9 for the other models).}
\label{fig:ebm}
\end{figure}

% [中文] 下图给出 Baichuan-EBM 上的循证医学结果（核心分 vs 引用精确率散点图）。
Figure~\ref{fig:ebm} reports the evidence-based medicine results on Baichuan-EBM.

% [中文] 图：表4（Baichuan-EBM 循证结果）数据的可视化——核心分 vs 引用精确率散点图（数据分析视角），公司橙色配色。
% [中文] Baichuan-M4 位于右上角，在两个维度上同时领先，引用精确率优势尤为明显（90 vs 43.8–55.9）。
% [中文] 图源脚本：make_ebm_chart.py（数据来自原 EBM 表）。

% [中文] 原始表格代码（已转为上方图片，保留以便回退）：
% \begin{table}[H]
% \centering\small
% \caption{Evidence-based medicine results on Baichuan-EBM (higher is better).}
% \label{tab:ebm}
% \begin{tabularx}{\linewidth}{@{}l CC@{}}
% \toprule
% \textbf{Model} & \textbf{Core Score $\uparrow$} & \textbf{Citation Precision $\uparrow$} \\
% \midrule
% \textbf{Baichuan-M4 (Ours)} & \textbf{83.1} & \textbf{90} \\
% GPT-5.5 & 79.2 & 54.7 \\
% Qwen-3.5 Plus & 60.4 & 43.8 \\
% OpenEvidence & 74.1 & 55.9 \\
% {\cjkfont 氢离子} & 71.5 & 52.6 \\
% \bottomrule
% \end{tabularx}
% \end{table}

% [中文] 5.3.2 医疗文档 OCR 评测
\subsubsection{Medical Document OCR Evaluation}

% [中文] 本评测衡量模型从版式复杂、含噪声的医疗文档中识别纯文本的能力，以及以结构化方式抽取核心字段的能力；同时报告高置信样本的覆盖量，以相似度阈值 >0.9 与 >0.8 界定。
% TODO-CITE: OCR 基线 PaddleOCR/GLM-OCR（可选，bib 暂无）
This evaluation measures the model's ability to recognize plain text from noisy medical documents with complex layouts, as well as its ability to extract core fields from structured inputs, where the outputs are later normalized into a structured format using an LLM for evaluation. The dataset contains a total of 297 samples. It also reports the coverage of high-confidence samples, defined by similarity thresholds of $>0.9$ and $>0.8$. Table~\ref{tab:ocr} reports the results.

% [中文] 表7：医疗文档 OCR 结果（越高越好）。样本数为高于相似度阈值的高置信样本数量。
% [中文] 表头：模型 | 非结构化分↑ | 结构化分↑ | 样本数 >0.9↑ | 样本数 >0.8↑
\begin{table}[H]
\centering\small
\caption{Medical document OCR results (higher is better). Sample counts are the number of high-confidence samples above the similarity thresholds.}
\label{tab:ocr}
\begin{tabularx}{\linewidth}{@{}l CCCC@{}}
\toprule
\textbf{Model} & \textbf{Unstructured Score $\uparrow$} & \textbf{Structured Score $\uparrow$} & \textbf{No. of Samples $>$0.9 $\uparrow$} & \textbf{No. of Samples $>$0.8 $\uparrow$} \\
\midrule
\textbf{Baichuan-M4 (Ours)} & \textbf{0.838} & \textbf{0.914} & \textbf{227} & \textbf{269} \\
Qwen3.5-397B-A17B~\cite{qwen3.5} & 0.644 & 0.871 & 167 & 245 \\
Qwen3.5-122B-A10B~\cite{qwen3.5} & 0.509 & 0.791 & 140 & 213 \\
PaddleOCR-VL-1.5~\cite{cui2026paddleocrvl15multitask09bvlm} & 0.329 & 0.777 & 117 & 196 \\
GLM-OCR~\cite{duan2026glmocrtechnicalreport} & 0.374 & 0.747 & 102 & 155 \\
\bottomrule
\end{tabularx}
\end{table}

% [中文] 5.3.3 X 光报告生成评测
\subsubsection{X-Ray Report Generation Evaluation}

This evaluation is conducted on the IU-Xray dataset~\cite{demner2016preparing}, a standard benchmark for chest X-ray report generation.

\begin{itemize}
\item \textbf{CIDEr}~\cite{vedantam2015cider}\textbf{:} Measures the readability of generated text and its language-level similarity to reference reports.
\item \textbf{GREEN-LLM}~\cite{ostmeier2024green}\textbf{:} Measures alignment with medical terminology and the clinical correctness of the underlying reasoning.
\end{itemize}

Table~\ref{tab:xray} presents the report generation results on the IU-Xray benchmark.

% [中文] 表8：X 光报告生成。
% [中文] 表头：模型 | CIDEr↑ | GREEN-LLM↑
\begin{table}[H]
\centering\small
\caption{Radiology report generation results on the IU-Xray benchmark (higher is better).}
\label{tab:xray}
\begin{tabularx}{\linewidth}{@{}l CC@{}}
\toprule
\textbf{Model} & \textbf{CIDEr $\uparrow$} & \textbf{GREEN-LLM $\uparrow$} \\
\midrule
\textbf{Baichuan-M4 (Ours)} & \textbf{0.1892} & \textbf{0.8435} \\
Gemini-3.1-Pro~\cite{gemini31pro2026} & 0.1593 & 0.8076 \\
Qwen3.5-122B-A10B~\cite{qwen3.5} & 0.1491 & 0.8348 \\
Qwen3.5-397B-A17B & 0.1305 & 0.8343 \\
Qwen3.5-35B-A3B~\cite{qwen3.5} & 0.1016 & 0.8305 \\
\bottomrule
\end{tabularx}
\end{table}

% [中文] 5.3.4 皮肤病专家系统评测（f17k）
\subsubsection{Dermatology Expert System Evaluation (f17k)}

% [中文] 本评测在包含 4893 例的 f17k 基准上进行。M4 引入了专为多轮视觉—临床交互设计的 Agent 架构。
This evaluation is conducted on a subset of the f17k benchmark~\cite{groh2021evaluating}. Specifically, 30\% of the cases were randomly sampled from the original dataset, and images without labels were subsequently removed, resulting in a final evaluation set of 4,893 cases. Although the f17k evaluation is image-only, it tests the diagnostic candidate-generation component of the dermatology agent. Table~\ref{tab:dermatology} reports the results. Multi-turn visual-clinical interaction remains to be evaluated in future work. 

Evaluation Metrics. Following clinical diagnosis practice, each model is required to generate a ranked list of six candidate diagnoses for each image. Predictions are mapped to ICD-11 codes and evaluated using hierarchical matching criteria. Specifically, \textbf{Exact}, \textbf{Category}, and \textbf{Block} correspond to matches at the 4-character, 3-character, and 2-character ICD-11 levels, respectively. We further report \textbf{TOP-3} and \textbf{TOP-6} accuracy, which indicate whether the ground-truth diagnosis is correctly identified within the top three or top six predictions.

% [中文] 表9：f17k 上的皮肤病专家系统结果（越高越好）。
% [中文] 表头：模型 | TOP-1 EXACT↑ | TOP-1 CATEGORY↑ | TOP-1 BLOCK↑ | TOP-3↑ | TOP-6↑
\begin{table}[H]
\centering\small
\caption{Dermatology expert system results on f17k (higher is better).}
\label{tab:dermatology}
\begin{tabularx}{\linewidth}{@{}l CCCCC@{}}
\toprule
\textbf{Model} & \textbf{TOP-1 EXACT $\uparrow$} & \textbf{TOP-1 CATEGORY $\uparrow$} & \textbf{TOP-1 BLOCK $\uparrow$} & \textbf{TOP-3 CATEGORY $\uparrow$} & \textbf{TOP-6 CATEGORY $\uparrow$} \\
\midrule
\textbf{Baichuan-M4 (Ours)} & \textbf{30.78\%} & \textbf{39.53\%} & \textbf{42.71\%} & \textbf{54.00\%} & \textbf{60.68\%} \\
Gemini-3.1-Pro & 27.55\% & 35.77\% & 39.71\% & 47.97\% & 55.20\% \\
Qwen3.5-397B-A17B & 25.10\% & 33.72\% & 37.34\% & 47.21\% & 55.39\% \\
Qwen3.5-9B~\cite{qwen3.5} & 18.64\% & 27.14\% & 30.51\% & 40.02\% & 48.07\% \\
\bottomrule
\end{tabularx}
\end{table}

% [中文] 5.4 关键结论总结
\subsection{Summary of Key Findings}

% [中文] 综合来看，工具调用与多模态评测主要体现出以下三点结论：
Taken together, the tool-use and multimodal evaluations point to three main findings:

\begin{itemize}
    % [中文] 高精度证据综合：在 Baichuan-EBM 检索测试中，M4 取得高达 90.0 的引用精确率，远超通用模型，表明其在使用外部工具时具备很强的去冗余与精准取证能力。
    \item \textbf{High-precision evidence synthesis.} In the Baichuan-EBM retrieval test, M4 achieves a very high Citation Precision score of 90.0, far ahead of general-purpose models. This indicates a strong capability in removing redundant information and extracting precise evidence when using external tools.
    % [中文] 端到端病历数字化：M4 突破了传统专用 OCR 软件的局限，在医疗文档上达到 0.914 的结构化字段抽取准确率，为下游医疗大数据分析构建了高质量的数据治理前端。
    \item \textbf{End-to-end medical record digitization.} M4 goes beyond the limits of traditional dedicated OCR software. It reaches a structured field extraction accuracy of 0.914 for medical documents, creating a high-quality data governance front end for downstream medical big data analysis.
    
    \item \textbf{Image perception and agent-based clinical reasoning.} On the IU-Xray benchmark, M4 achieves the strongest report generation performance among the compared multimodal models, demonstrating its image understanding and medical report generation capabilities. In complex dermatology scenarios that heavily rely on ``visual examination'' M4 further leverages agentic multi-step differential diagnosis reasoning, achieving the best TOP-1 exact disease match rate (30.78\%) and TOP-6 diagnosis inclusion rate (60.68\%).

\end{itemize}

% [中文] 六、结论
\section{Conclusion}

% [中文] Baichuan-M4 是一款面向连续照护场景的临床级医疗大模型，不再局限于单轮医学问答，而是通过 Baichuan-Harness、核心模型与工具系统组成医疗 Agent 体系，支持问诊、随访、慢病管理、循证检索、医学影像理解和长期患者记忆管理等任务。其核心升级包括多 Agent 协作、动态角色切换、长期患者记忆、安全护栏、闭环强化学习训练，以及面向病历、文献、影像和皮肤病等场景的专业工具调用能力。同时，M4 强调可追溯的循证医学能力和多模态医学感知能力，在医疗文档 OCR、X 光分析、皮肤病诊断辅助等方面形成了较完整的临床工具链。
Baichuan-M4 is a clinical-grade medical agent system designed for physician-supervised continuous care. Rather than answering one-off medical questions, it works as a medical agent system that combines Baichuan-Harness, a core model, and a set of clinical tools to handle consultation, follow-up, chronic disease management, evidence retrieval, medical image understanding, and long-term patient memory. Its main upgrades are multi-agent coordination, dynamic role switching, persistent patient memory, safety guardrails, and closed-loop reinforcement-learning training, together with tool use tailored to medical records, literature, imaging, and dermatology. M4 also puts weight on traceable evidence-based reasoning and multimodal perception, pulling document OCR, X-ray analysis, and dermatology support into a fairly complete clinical toolchain.

% [中文] 评测结果显示，Baichuan-M4 在复杂临床推理、安全幻觉控制、长上下文医学记忆、证据检索精度和多模态医学理解等方面均有明显提升，体现出其作为严肃医疗场景辅助系统的应用潜力，但其输出仍应作为医生和医疗机构的辅助参考，不能替代执业医生的最终诊断与治疗决策。
Across our benchmarks, M4 improves clearly on complex clinical reasoning, hallucination control, long-context clinical memory, retrieval precision, and multimodal understanding, which points to its value as an assistant in serious medical settings. Still, its outputs are meant as a reference for physicians and healthcare institutions, and should not replace a licensed physician's final diagnosis or treatment decisions.

% [中文] 七、贡献者
\section{Contribution}\label{sec:contribution}

% [中文] 贡献者按名字（first name）的字母顺序排列。星号（*）表示已不在团队中的成员。
Contributors are presented in alphabetical order according to their first names. An asterisk (*) denotes those who are no longer part of the team. A dagger (\textsuperscript{$\dagger$}) indicates special acknowledgment. We gratefully thank Professor Lijie Wen from the School of Software, Tsinghua University, for his valuable advice and support.

% [中文] 核心贡献者
\subsection*{Contributors}

Aiyuan Yang,
Chengfeng Dou,
Da Pan,
Dian Wang,
Fan Yang*,
Fei Deng,
Fei Li,
Guangwei Ai,
Hui Liu,
Hongda Zhang,
Jinyang Tai,
Jiyuan Jia,
Kai Lu,
Lijun Liu,
Linwei Chen,
Linyu Li,
Meiqing Guo,
Peidong Guo,
Qiang Ju,
Rihui Xin,
Shuai Wang,
XinKai Ma,
Yichuan Mo,
Yijie Zhou,
Canbin Piao\textsuperscript{$\dagger$},
Leyi Pan\textsuperscript{$\dagger$},
Yihe Luo\textsuperscript{$\dagger$},
Zian Wang\textsuperscript{$\dagger$}

\subsection*{Experts and Advisors}

Hengfu Cui*,
Zhishou Zhang,
Yifan Zhang,
Meng Tian,
Huijing Wang,
Ziqiang Zeng,
Yawen Wang,
Yan Zhang,
Lijie Wen\textsuperscript{$\dagger$}

% [中文] 参考文献（References）：仅显示被 \cite 引用的条目；数字 [1] 样式由 biblatex 自动编号。
\printbibliography[title={References}]

\end{document}